\pdfoutput=1

\documentclass[sigconf,natbib=false, nonacm]{acmart}
\AtBeginDocument{%
  }

\setcopyright{acmlicensed}
\copyrightyear{2025}
\acmYear{2025}
\acmDOI{XXXXXXX.XXXXXXX}
\acmConference[ICAIL '25]{The 20th International Conference on Artificial Intelligence and Law}{June 16--20,
  2025}{Chicago, IL}

\RequirePackage[
  datamodel=acmdatamodel,
  style=acmnumeric,
  ]{biblatex}
\usepackage{subcaption}

\newcommand\blfootnote[1]{%
  \begingroup
  \renewcommand\thefootnote{}\footnote{#1}%
  \addtocounter{footnote}{-1}%
  \endgroup
}

\begin{document}

\title{Parameterized Argumentation-based Reasoning Tasks for Benchmarking Generative Language Models}


\author{Cor Steging}
\email{c.c.steging@rug.nl}
\orcid{0000-0001-6887-1687}
\affiliation{%
  \institution{Bernoulli Institute of Mathematics, Computer Science and Artificial Intelligence, University of Groningen}
}

\author{Silja Renooij}
\email{s.renooij@uu.nl}
\orcid{0000-0003-4339-8146}
\affiliation{%
  \institution{Department of Information and Computing Sciences,\\ Utrecht University}
}

\author{Bart Verheij}
\email{bart.verheij@rug.nl}
\orcid{0000-0001-8927-8751}
\affiliation{%
  \institution{Bernoulli Institute of Mathematics, Computer Science and Artificial Intelligence, University of Groningen}
}

\renewcommand{\shortauthors}{Steging et al.}

\begin{abstract}
Generative large language models as tools in the legal domain have the potential to improve the justice system. 
However, the reasoning behavior of current generative models is brittle and poorly understood, hence cannot be responsibly applied in the domains of law and evidence. 
In this paper, we introduce an approach for creating benchmarks that can be used to evaluate the reasoning capabilities of generative language models. 
These benchmarks are dynamically varied, scalable in their complexity, and have formally unambiguous interpretations.
In this study, we illustrate the approach on the basis of witness testimony, focusing on the underlying argument attack structure. 
We dynamically generate both linear and non-linear argument attack graphs of varying complexity and translate these into reasoning puzzles about witness testimony expressed in natural language. 
We show that state-of-the-art large language models often fail in these reasoning puzzles, already at low complexity.
Obvious mistakes are made by the models, and their inconsistent performance indicates that their reasoning capabilities are brittle.
Furthermore, at higher complexity, even state-of-the-art models specifically presented for reasoning capabilities make mistakes. 
We show the viability of using a parametrized benchmark with varying complexity to evaluate the reasoning capabilities of generative language models. 
As such, the findings contribute to a better understanding of the limitations of the reasoning capabilities of generative models, which is essential when designing responsible AI systems in the legal domain. 
\end{abstract}



\maketitle
\blfootnote{This manuscript has been accepted for presentation as a short paper at the 20th International Conference of AI \& Law in Chicago, June 16 to 20 of 2025.}

\section{Introduction}
Generative large language models have shown potential in the legal domain, for example by annotating legal texts~\cite{Savelka2023TheUnreasonable}, making legal information more accessible to laypeople~\cite{chien2024generative}, or by improving the usability of court forms~\cite{steenhuis2023beyond}.
Generative language models are, however, not without flaws. One particular flaw of generative models is that they can produce hallucinations, which can be described as unfaithful or nonsensical output~\cite{hallucinations2023ji}. 
One study suggests that in legal settings, popular generative language models hallucinate at least 58\% of the time~\cite{Dahl2024LargeLegalFictions}.
Apart from hallucinations in language models, it has been shown that, more generally, data-driven machine learning methods can produce the right legal conclusion as output while using the wrong reasoning~\cite{StegingICAIL21}. In the legal domain, correct decisions must be made and these right decisions should be backed up by sound arguments. In other words, models should use sound legal reasoning to come to a decision. 
Two approaches have been introduced to deal with this lack of correct reasoning. 
The first is to create hybrid systems that combine generative language models with logical formalisms~\cite{kant2024equitableaccessjusticelogical, StegingJURISIN2024, westermann2024dallma}.
The second is to adapt the generative language models specifically for reasoning tasks, either by fine-tuning the model itself~\cite{Liga2023Finetuning}, or by making use of the model's in-context learning through prompt engineering methods, such as chain-of-thought prompting~\cite{wei2022chain}, in order to elicit the correct response.
While there have been initial studies documenting emerging reasoning capabilities in generative language models, these capabilities are still not well understood~\cite{Emergent2023Webb, huang-chang-2023-towards}.

For evaluating the legal reasoning capabilities of generative language models, a number of benchmarks have been introduced to test the models on various legal reasoning tasks~\cite{legalbench}. 
These benchmarks are datasets that usually consist of a fixed set of questions and the correct answers to these questions. 
Performance is then measured by how well the models can answer the questions correctly.
While this does not investigate the reasoning patterns of the models directly, it does provide insight into the extent to which the models can solve reasoning tasks.
An issue with this type of benchmark is that they are static and hence allow for memorizing effects by data contamination (in the sense of mixing test and training data). This is significant, since once they are published, commercially available and other generative language models can be (and often are) adjusted or retrained using the specific questions from the benchmark, which increases the models' performance on the benchmark.
Still, the model might perform poorly on different but similar legal reasoning tasks that the model was not trained on explicitly. 
This cat-and-mouse game makes it difficult to assess to what extent generative models have been able to generalize proper reasoning behavior on the basis of their training, typically using immense amounts of unclearly specified data. 

In this paper, we therefore propose to design legal reasoning benchmarks that can be generated dynamically, reducing the issue of data contamination. Our approach furthermore allows for parametrized scalability in terms of problem complexity, such that models may be tested to their limits. 
More specifically, we propose an approach where we generate argument attack graphs and automatically translate these into natural language prompts. 
In the upcoming sections, we describe how these attack graphs are generated and how they are translated into prompts for generative language models. 
Furthermore, we illustrate our approach by generating three types of benchmarks of increasing difficulty, based on linear attack graphs, non-linear attack graphs, and non-linear attack graphs with shuffled arguments in the prompt. We demonstrate the viability of our approach by using these three benchmarks to evaluate the reasoning performance of seven contemporary generative language models: GPT-4o-mini, GPT-4o, o1-preview, Gemini-1.5-flash, Gemini-1.5-pro, Claude-3.5-haiku, and Claude 3.5-sonnet. 

\section{Generating parameterized benchmarks}
For the benchmarks that we generate, we aim for three properties. First, the benchmark should be dynamically varied as a measure against the problem of data contamination. Second, the complexity of the problems should be parametrized to allow for scalability, in order to test the limitations of state-of-the-art models. Third, the problems should have a formally unambiguous interpretation.

We therefore propose an approach wherein we generate formal reasoning tasks of parameterized scaling complexity that have formally unambiguous interpretations, and we translate these tasks into natural language using an ontology to create variations in order to avert data contamination. 

As a formal background, we choose argument attack as studied in artificial intelligence~\cite{dung1995,baroniEtal2020}. Specifically, we use argument attack graphs (in the literature also referred to as abstract argumentation frameworks~\cite{dung1995}) $(\mathcal{A}, \mathcal{E})$ in which the nodes $\mathcal{A} = \{A_1,\ldots, A_n\}$ represent a set of $n$ arguments $A_i$ and each directed edge $A_i \leftarrow A_j$ in $\mathcal{E}$ represents an argument attack, where argument $A_j$ attacks argument $A_i$. To illustrate our approach, two types of attack graphs are investigated in this paper: strictly linear attack graphs and non-linear attack graphs. 

We have selected the legally relevant topic of witness testimony, where reasoning puzzles presented to the language models are based on the idea that witnesses can in principle be believed unless there is evidence (from another witness) that they may be lying. 
In this study, each node in an argument attack graph represents a one-step argument from a witness testimony to the claim made by the witness, and each edge from node $A$ to node $B$ represents that the argument represented by $B$ is attacked by the argument of $A$ since the witness testimony in $A$ claims that the witness of $B$ lies.

\begin{figure}
\begin{subfigure}[b]{\columnwidth}
    \centering
    \includegraphics[width=0.25\linewidth]{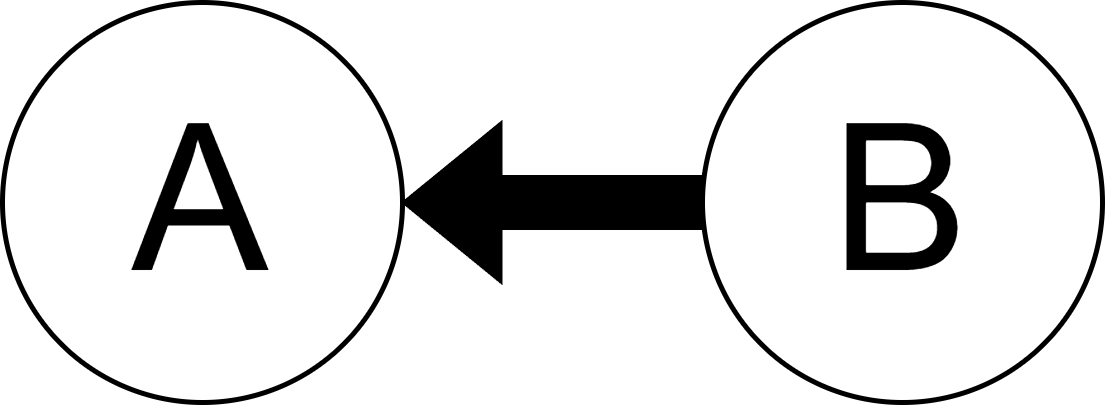}
    \caption{Linear argument attack graph with two arguments}
    \label{fig:example_linear_graph}
\end{subfigure}

\par\bigskip 
\begin{subfigure}[b]{\columnwidth}
\raggedright
    \footnotesize
    \texttt{The following is a reasoning puzzle. 
     Witnesses should be believed unless there is testimony that they are lying. Now consider the following facts:}
     \newline \newline
     \texttt{Witness Alice says that the train is late. \\
     Witness Bob says that witness Alice is lying. 
     \newline \newline
     Question: should it be believed that the train is late? \\
     End your answer with: "Answer: yes or no".}
    \caption{Example of a generated prompt based on the graph in Figure~\ref{fig:example_linear_graph}}
    \label{fig:example_prompt}    
\end{subfigure}
\caption{Example of a linear argument attack graph and generated prompt}
\end{figure}

\subsection{Linear attack graphs}
We define a linear attack graph $(\mathcal{A},\mathcal{E})$ as a graph consisting of a single directed path $A_1 \leftarrow \ldots \leftarrow A_n$, in which every argument $A_i \in \mathcal{A}$, $1 \leq i < n$, is attacked by $A_{i+1}$.
An example of such a linear attack graph with $\mathcal{A}=\{A,B\}$ can be seen in Figure~\ref{fig:example_linear_graph}. Arguments are evaluated using the basic principle `The one who has the last word laughs best'~\cite{dung1995}. Concretely, if argument $A_1$ is the only argument, it is accepted since it `has the last word'. If it is followed by an attacking argument $A_2$, that argument $A_2$ is the accepted argument having the last word, and $A_1$ is rejected. If in turn $A_2$ is attacked by $A_3$, the argument $A_3$ is accepted, $A_2$ is rejected, and also $A_1$ is accepted since it is defended by $A_3$ against the attack by $A_2$. The argument $A_1$ is then said to be reinstated.

\begin{figure}
\begin{subfigure}[b]{\columnwidth}
    \centering
    \includegraphics[width=0.375\linewidth]{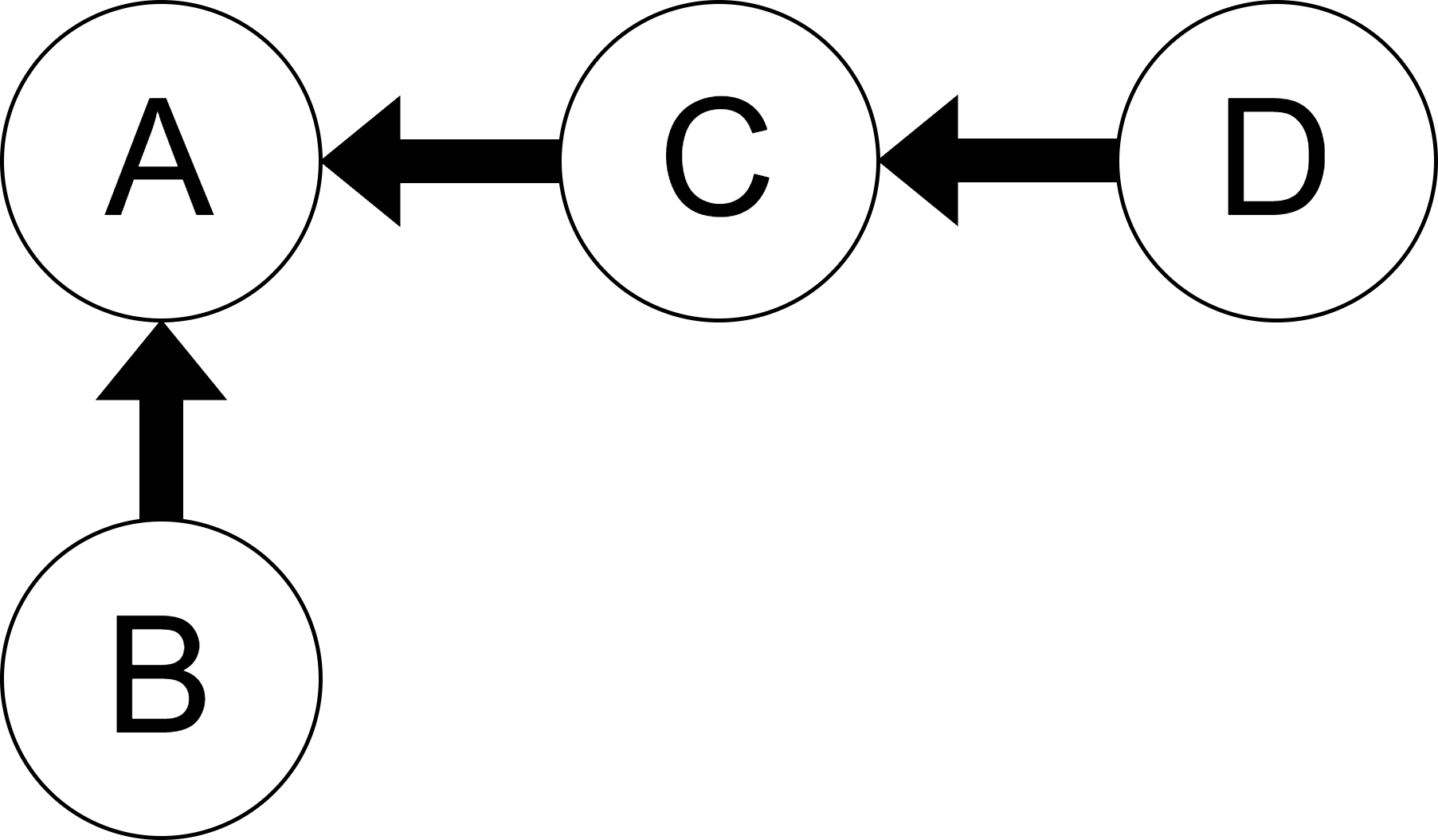}
    \caption{A non-linear attack graph with four arguments}
    \label{fig:example_nonlinear_graph}
\end{subfigure}
\par\bigskip
\begin{subfigure}[b]{\columnwidth}
\raggedright
    \footnotesize
    \texttt{The following is a reasoning puzzle. 
     Witnesses should be believed unless there is testimony that they are lying. Now consider the following facts:}
     \newline \newline
     \texttt{Witness Alice says that the train is late. \\
     Witness Bob says that witness Alice is lying. \\
     Witness Charlie says that witness Alice is lying. \\
     Witness Dan says that witness Charlie is lying.}
     \newline \newline
     \texttt{Question: should it be believed that the train is late? \\
     End your answer with: "Answer: yes or no".}
    \caption{Example of a generated prompt based on the graph in Figure~\ref{fig:example_nonlinear_graph}}
    \label{fig:example_prompt_nonlinear}
\end{subfigure}
\caption{Example of a non-linear argument attack graph and an generated prompt}
\end{figure}

More generally, linear attack graphs have a straightforward formal interpretation in the approach by~\cite{dung1995}, as follows: if a directed path of arguments $A_1 \leftarrow \ldots \leftarrow A_n$ 
has an even length $n$, the arguments with an even index $i$ are accepted, and those with an odd index rejected, while if the length $n$ is odd, the odd-indexed arguments are accepted, and the even ones rejected. In~\cite{dung1995}'s terminology, for our linear graphs, the grounded, stable, and preferred semantics all lead to the same, unambiguous outcome. Our focus in the benchmarks is on the first argument $A_1$, which is accepted when $n$ is odd, and rejected when $n$ is even. 
As our linear attack graphs are a directed path, only one unique linear attack graph can exist for any total number of arguments $n$.

\subsection{Non-linear attack graphs}
\label{sec:non-linear}
In addition to linear attack graphs, where each argument attacks the previous argument, we also generate non-linear attack graphs. In this study, we create non-linear attack graphs where multiple arguments $C_{i1}$ attack the first argument $A_1$, and each $C_{i1}$ is the first argument in a directed path of arguments. More specifically, we define a non-linear attack graph $(\mathcal{A},\mathcal{E})$ as a graph with $n$ arguments $\mathcal{A} = \{A_1\} \cup \bigcup_{i=1}^{k} \mathcal{C}_i$, $\bigcap_{i=1}^{k} \mathcal{C}_i = \emptyset$, and edges $\mathcal{E} = \{A_1 \leftarrow C_{i1} \mid C_{i1} \in \mathcal{C}_i, 1\leq i\leq k \} \cup \bigcup_{i=1}^{k} \mathcal{E}_i$ such that each of the $k$ subgraphs $(\mathcal{C}_i,\mathcal{E}_i)$ is a disjoint linear attack graph with first argument $C_{i1}$.
An example of such a non-linear attack graph can be seen in Figure~\ref{fig:example_nonlinear_graph}, where the main argument $A$ is attacked by two arguments: $B$ and $C$. Argument $C$ in turn is attacked by argument $D$. Both $B$ and $C \leftarrow D$ are directed paths as in the definition ($k=2$), with lengths 1 and 2 respectively. Since $D$ attacks $C$, we cannot accept $C$, so $C$'s attack of $A$ fails. However, because argument $B$ also attacks argument $A$, argument $A$ cannot be accepted. 

In the non-linear attack graphs that we generate, the first argument $A_1$ is accepted if and only if all directed paths attached to $A_1$ are of even length. 
That is, the main argument $A_1$ can be accepted if and only if $\vert \mathcal{C}_i\vert$ is even for all $i=1,\ldots,k$; otherwise $A_1$ is rejected.

A given number of arguments $n > 2$ corresponds to multiple, differently structured graphs, one for each integer partition of $n-1$. This number of different graph structures increases exponentially with the number of arguments $n$ in the graph.

\subsection{Ontology-based prompt generation}
To evaluate whether generative language models have the reasoning capabilities to solve for the attack graphs we generate, the graphs need to be presented to the model as a prompt in natural language. 
In our approach, we automatically convert the attack graphs into natural language with the use of an ontology that represents varied context information, and we randomly select specific elements of the ontology to be used in the arguments of the attack graph.
That way, we can vary the content of each argument and thus make diverse prompts for each attack graph. 
In this study, the ontology is a list of names and witness statements.

The first argument of every graph is a statement made by a witness. All subsequent arguments are testimony from other witnesses that claim that some witness lies, representing an attack. The question posed at the end of the prompt is whether the statement made by the first witness should be believed given all of the arguments. 
By using a list of 474 unique names and 90 unique statements, we generate a wide variation of prompts. 

An example of a prompt for a linear attack graph with two arguments can be seen in Figure~\ref{fig:example_prompt}, based on the graph in Figure~\ref{fig:example_linear_graph}. Note that we add some additional information and instructions to aid the generative language model in providing the correct output. When generating the prompts, we also store the correct answer to easily evaluate the models' performance. For the example prompt in Figure~\ref{fig:example_prompt}, the correct answer should be `no' as the argument using Alice's testimony is attacked by the argument using Bob's, thus we should not believe what Alice has said. For linear attack graphs, the correct answer for graphs with an odd number of arguments is always `yes' whereas the correct answer for graphs with an even number of arguments is always `no'.
Prompts for non-linear attack graphs are generated in a similar fashion. In Figure~\ref{fig:example_prompt_nonlinear}, we show an example of a prompt that could be generated from the graph shown in Figure~\ref{fig:example_nonlinear_graph}. For non-linear attack graphs, the correct answer is `yes' if and only if all directed paths attached to the main argument have an even length, otherwise the answer is `no'.

We vary the default names of `Alice', `Bob', `Charlie' and `Dan' and statement `the train is late' using a basic ontology that consists of a list of other names (such as `Landry', `Ellie', `Nikolas' and `Avani') and statements (`the mirror is broken', `the sky is cloudy'). We avoid the use of nonsensical statements that might confuse the model, such as `the grass is purple', or statements that are always true or always false (`the water is wet', 'the square is a circle').  

\subsection{Overview}
\begin{figure*}
    \centering
    \includegraphics[width=0.6\linewidth]{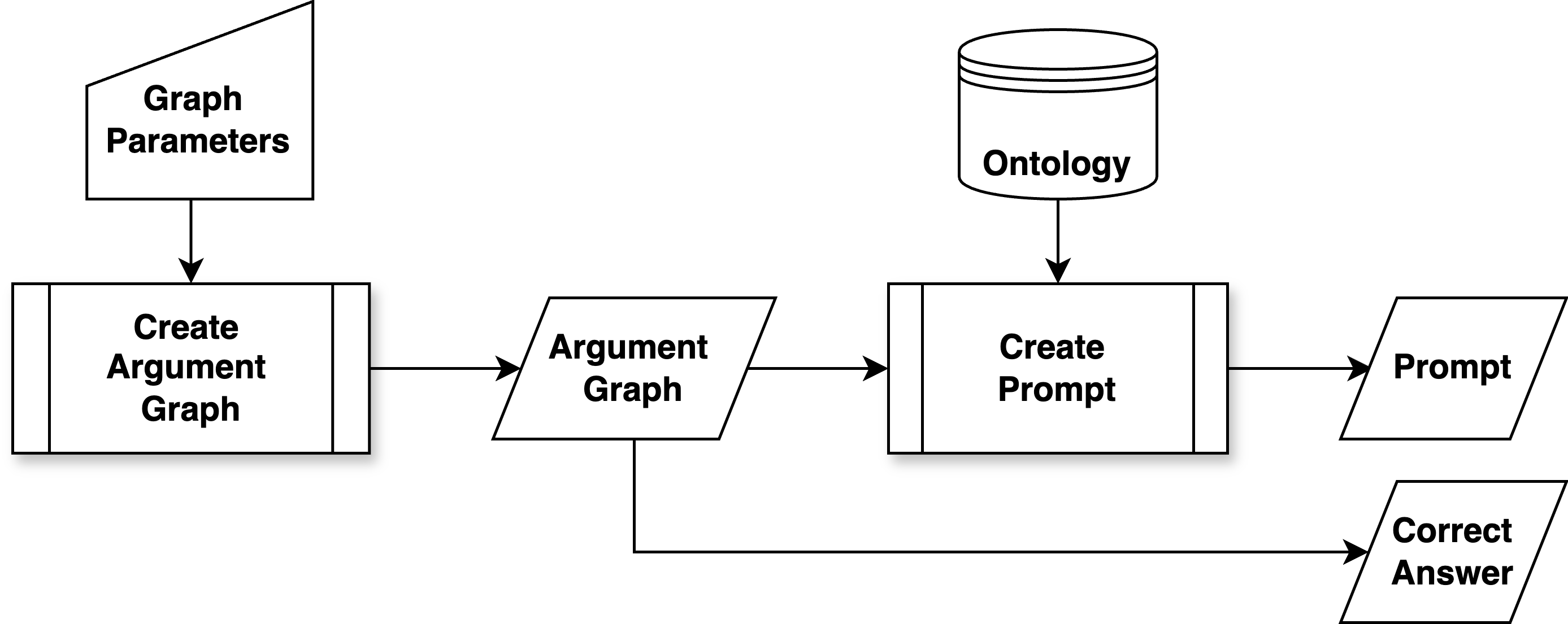}
    \caption{The pipeline of our approach for generating dynamic benchmarks of scaling complexity. In this paper, the graph parameters are the number of arguments in the graph, and the ontology is a list of names and statements}
    \label{fig:pipeline}
\end{figure*}

The complete pipeline of our approach to creating benchmarks can be seen in Figure~\ref{fig:pipeline}. Following this pipeline, we first generate an attack graph based on a set of parameters. For the linear graphs, this is only the total number $n$ of arguments that should be in the graph. For the non-linear graphs, this is the number $k$ of directed paths in the graph and the total number of arguments in each directed path. Once a graph has been generated, we convert that graph into a prompt using an ontology, which in this study is a list of names and statements. The resulting prompt can then be provided to a generative language model and the model's output can be compared to the correct answer to determine the performance of the model.

Our setup meets the three properties we started with as follows. For \textit{dynamic variation}, we use the context ontology information, here a list of statements and names. For \textit{scalable complexity}, we use the graph parameters of the argument attack graphs. For \textit{formally unambiguous interpretation}, we use 
attack graphs with a formally unambiguous interpretation in formal argumentation semantics.

\section{Method}
To illustrate our approach, we perform a set of experiments where we generate benchmarks using the approach described in the previous section, and use these benchmarks to evaluate the reasoning capabilities of generative language models. We evaluate the models using generated prompts based on linear, and non-linear attack graphs. Additionally, to potentially increase the difficulty of the task further, we perform an additional experiment where we shuffle order of the arguments in the non-linear prompts. The performance of the models on these prompts is evaluated and compared to their performance on the non-shuffled non-linear prompts.

\subsection{Models}
We evaluate a total of seven popular, commercially available generative language models using our approach. An overview of the models that we evaluate can be seen in Table~\ref{tbl:models_used}. We opted to evaluate the latest models from OpenAI\footnote{\url{https://platform.openai.com/docs/models}, accessed 14 January 2025}, Google\footnote{\url{https://ai.google.dev/gemini-api/docs/models/gemini}, accessed 14 January 2025}, and Anthropic\footnote{\url{https://docs.anthropic.com/en/docs/about-claude/models}, accessed 14 January 2025}, and assess both their flagship model, as well as their smaller, lightweight model. Additionally, we evaluate OpenAI's o1-preview model, a preview model specifically presented for its reasoning capabilities.

\begin{table}[]
\caption{Overview of the models used in our experiments}
\begin{tabular}{r|r|l}
\textbf{Company}   & \textbf{Model}    & \textbf{Summary} \\ \hline
\textbf{OpenAI}    & GPT-4o-mini       & Fast model       \\
                   & GPT-4o            & Flagship model   \\
                   & o1-preview        & Reasoning model  \\\hline
\textbf{Google}    & Gemini-1.5-flash  & Fast model       \\
                   & Gemini-1.5-pro    & Flagship model   \\\hline
\textbf{Anthropic} & Claude-3.5-haiku  & Fast model       \\
                   & Claude-3.5-sonnet & Flagship model  
\end{tabular}
\label{tbl:models_used}
\end{table}

\subsection{Benchmark datasets}
For the experiment concerning linear attack graphs, we generate linear graphs with 1 to 25 arguments (and in one case to 50). For each of these 25 (or 50) graphs, we generate 100 unique prompts using the ontology, yielding a total of 2,500 (or 5,000) prompts. By design, the correct answer is `yes' in 52\% of the 2,500 (or 48\% of the 5,000) prompts. This is due to the fact that the answer is `yes' for all odd number of arguments and `no' otherwise. Each of the seven language models is evaluated on the same prompts. We evaluate the models by comparing their output to the correct answer for each prompt. For the o1-preview model, which is still in preview, we only run the first 75 prompts (three prompts for each number of arguments), as the costs of running this experimental model are significantly higher. For all models, we expect performance to go down for prompts with a larger number of arguments.

We perform a similar experiment with non-linear attack graphs, where we generate all unique non-linear attack graphs with 1 to 15 arguments, yielding a total of 508 unique graphs. Specifically, we generate each unique attack graph for a given total number of arguments $n$ by generating all possible partitions for $n-1$. 
Each unique partition will determine the number and lengths of the directed path that are attached to the main argument. For instance, for $n=5$, we generate all partitions for the number $n-1=4$, which has 5 possible partitions ([4], [3, 1], [2,2], [2,1,1], [1,1,1,1]), and thus 5 unique graphs in our setup, namely: 
the graph with a single directed path of 4 arguments attacking the main argument (giving a linear attack graph of length 5), the graph with one directed path of length 3 and one of 1, the graph with two directed paths of length 2, the graph with three directed paths of lengths 2, 1 and 1, respectively, and the graph with four directed paths of 1 argument.
For a complete overview of the graph generation process, we refer to the code repository.\footnote{Code and datasets can be found at \url{https://github.com/CorSteging/ParameterizedArgumentationBasedReasoningTasks/}}

For each of the 508 unique graphs, we generate 5 unique prompts using the ontology, which yields a total of 2,540 prompts based on non-linear attack graphs. The correct answer is `yes' in only 8.85\% of the 2,540 non-linear prompts. We evaluate each of the seven language models by presenting them with the prompts, and comparing their output to the correct answers. We choose to evaluate the o1-preview model using only the hardest prompts, as this model is more costly. We select these hard prompts by collecting the prompts that were answered incorrectly by the best performing models. We expect the non-linear prompts to be more difficult, and therefore that the performance of the models will be lower on these prompts than on the strictly linear prompts. 

By default, all of the arguments in the prompt are ordered based on the attack relationships, where the first argument is attacked by the second, and so forth. As an additional experiment, we investigate whether shuffling the order of the arguments in the prompt makes a difference. We perform this additional experiment by shuffling all 2,540 prompts that were generated from non-linear attack graphs as mentioned earlier. We then provide these shuffled prompts to the model and compare the performances of the models to their performances on original, non-shuffled prompts. We expect this shuffling to make the task more difficult for the models.

\begin{table*}[]
\caption{The accuracy, F1-score, Matthew's Correlation Coefficient (MCC), recall and precisions of the models for prompts generated from linear attack graphs (\ref{tbl:linear_results}), non-linear attack graphs (\ref{tbl:nonlinear_results}), and non-linear attack graphs with shuffled statements (\ref{tbl:nonlinear_shuffled_results}). Furthermore, the number of arguments (args), number of prompt variations (vars), total number of generated prompts, and number of parsed prompts for each experiment. Best performances are shown in bold (excluding o1-preview performance)}
\begin{subtable}[h]{\textwidth}
\caption{Linear graphs}
\begin{tabular}{r|cccc|ccccc}
   & \multicolumn{4}{c|}{\textbf{Experimental settings}} & \multicolumn{5}{c}{\textbf{Model Performance}}                                                \\
\textbf{Model}                    & \textbf{\#Args}  & \textbf{\#Vars} & \textbf{\#Prompts} & \textbf{\#Parsed} & \textbf{Accuracy} & \textbf{F1-score} & \textbf{MCC}   & \textbf{Recall} & \textbf{Precision} \\ \hline
\textbf{GPT-4o-mini}              & 1-25                  & 100                    & 2,500                & 2,499               & 54.38             & 57.3              & 8.42           & 55.84           & 58.85              \\
\textbf{GPT-4o}                   & 1-25                  & 100                    & 2,500                & 2,499               & 72.31             & 77.58             & 47.52          & 67.02           & 92.08              \\
\textbf{Gemini-1.5-flash}         & 1-25                  & 100                    & 2,500                & 2,500               & 64.64             & 70.26             & 29.7           & 62.44           & 80.31              \\
\textbf{Gemini-1.5-pro}           & 1-25                  & 100                    & 2,500                & 2,500               & 60.32             & 60.03             & 20.91          & 63.03           & 57.31              \\
\textbf{Claude-3-5-haiku}  & 1-25                  & 100                    & 2,500                & 2,500               & 58.84             & 69.16             & 19.53          & 56.65           & 88.77              \\
\textbf{Claude-3-5-sonnet} & 1-25                  & 100                    & 2,500                & 2,500               & \textbf{77.32}    & \textbf{80.89}    & \textbf{56.59} & \textbf{71.99}  & \textbf{92.31}     \\ \hline
\textbf{o1-preview}               & 1-25                  & 3                      & 75                  & 75                 & 100.0             & 100.0             & 100.0          & 100.0           & 100.0         
    \\ \hline
\textbf{GPT-4o}                   & 1-50                  & 100                    & 5,000              & 4,998               & 63.05             & 70.33             & 29.92          & 58.77           & 87.56              
\end{tabular}
\label{tbl:linear_results}
\end{subtable}

\vspace*{0.5 cm}

\begin{subtable}[h]{\textwidth}
\caption{Non-linear graphs}
\begin{tabular}{r|cccc|ccccc}
   & \multicolumn{4}{c|}{\textbf{Experimental settings}} & \multicolumn{5}{c}{\textbf{Model Performance}}                                                \\
\textbf{Model}                    & \textbf{\#Args}  & \textbf{\#Vars} & \textbf{\#Prompts} & \textbf{\#Parsed} & \textbf{Accuracy} & \textbf{F1-score} & \textbf{MCC}   & \textbf{Recall} & \textbf{Precision} \\ \hline
\textbf{GPT-4o-mini}              & 1-15                  & 5                      & 2540                 & 2540               & 74.57             & 28.22             & 21.08          & 18.81           & 56.44              \\
\textbf{GPT-4o}                   & 1-15                  & 5                      & 2540                 & 2538               & 51.77             & 25.82             & 24.21          & 14.95           & \textbf{94.67}     \\
\textbf{Gemini-1.5-flash}         & 1-15                  & 5                      & 2540                 & 2540               & 46.46             & 21.02             & 13.63          & 12.09           & 80.44              \\
\textbf{Gemini-1.5-pro}           & 1-15                  & 5                      & 2540                 & 2540               & 58.46             & 23.72             & 17.09          & 14.16           & 72.89              \\
\textbf{Claude-3-5-haiku}  & 1-15                  & 5                      & 2540                 & 2533               & 33.6              & 18.9              & 9.93           & 10.6            & 87.11              \\
\textbf{Claude-3-5-sonnet} & 1-15                  & 5                      & 2540                 & 2540               & \textbf{86.81}    & \textbf{51.38}    & \textbf{48.73} & \textbf{38.15}  & 78.67            
\end{tabular}
\label{tbl:nonlinear_results}
\end{subtable}

\vspace*{0.5 cm}

\begin{subtable}[h]{\textwidth}
\caption{Non-linear graphs with shuffled arguments}
\begin{tabular}{r|cccc|ccccc}
   & \multicolumn{4}{c|}{\textbf{Experimental settings}} & \multicolumn{5}{c}{\textbf{Model Performance}}                                                \\
\textbf{Model}                    & \textbf{\#Args}  & \textbf{\#Vars} & \textbf{\#Prompts} & \textbf{\#Parsed} & \textbf{Accuracy} & \textbf{F1-score} & \textbf{MCC}   & \textbf{Recall} & \textbf{Precision} \\ \hline
\textbf{GPT-4o-mini}              & 1-15                  & 5                      & 2540                & 2540               & 83.35             & 29.38             & 21.46          & 39.11              & 23.53           \\
\textbf{GPT-4o}                   & 1-15                  & 5                      & 2540                & 2534               & 75.41             & 35.57             & \textbf{32.35} & \textbf{76.44}     & 23.18           \\
\textbf{Gemini-1.5-flash}         & 1-15                  & 5                      & 2540                & 2540               & 75.83             & 27.25             & 19.42          & 51.11              & 18.58           \\
\textbf{Gemini-1.5-pro}           & 1-15                  & 5                      & 2540                & 2540               & 79.06             & 26.11             & 17.54          & 41.78              & 18.99           \\
\textbf{Claude-3-5-haiku}  & 1-15                  & 5                      & 2540                & 2521               & 61.8              & 18.46             & 6.98           & 48.88              & 11.38           \\
\textbf{Claude-3-5-sonnet} & 1-15                  & 5                      & 2540                & 2540               & \textbf{89.72}    & \textbf{37.71}    & 32.25          & 35.11              & \textbf{40.72}  \\
\end{tabular}
\label{tbl:nonlinear_shuffled_results}
\end{subtable}

\vspace*{0.5 cm}

\begin{subtable}[h]{\textwidth}
\caption{Performance of o1-preview on a selection of `hard' non-linear prompts}
\begin{tabular}{r|rccc|ccccc}
   & \multicolumn{4}{c|}{\textbf{Experimental settings}} & \multicolumn{5}{c}{\textbf{Model Performance}}                                                \\
\textbf{Model} & \textbf{\#Args} & \textbf{\#Vars} & \textbf{\#prompts} & \textbf{\#Parsed} & \textbf{Accuracy} & \textbf{F1-score} & \textbf{MCC} & \textbf{Recall} & \textbf{Precision}  \\ \hline
\hspace{0.9cm} \textbf{o1-preview} & 4-15  & 1 & 262 & 262 & 91.6              & 52.17             & 56.73        & 35.29              & 100.0    \\           \end{tabular}
\label{tbl:nonlinear_o1}
\end{subtable}
\label{tbl:all_results}
\end{table*}

\begin{figure}
    \centering
    \includegraphics[width=\linewidth]{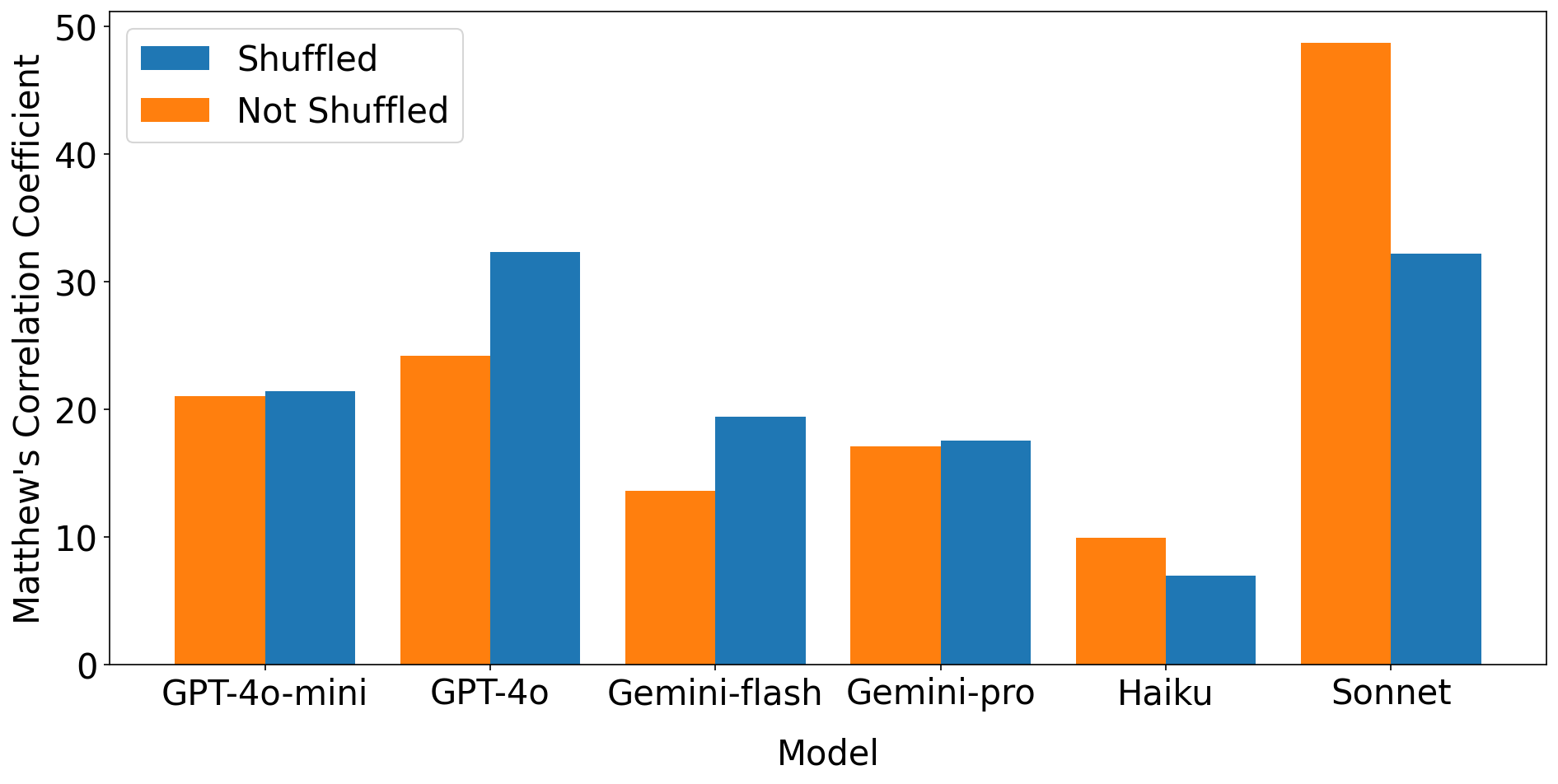}
    \caption{The mean MCC of models on prompts that are shuffled and not shuffled based on non-linear attack graphs}
    \label{fig:MCC_shuffled_diagram}
\end{figure}

\begin{figure*}[h!]
     \centering
     \begin{subfigure}[b]{0.49\textwidth}
         \centering
        \includegraphics[width = 1 \linewidth]{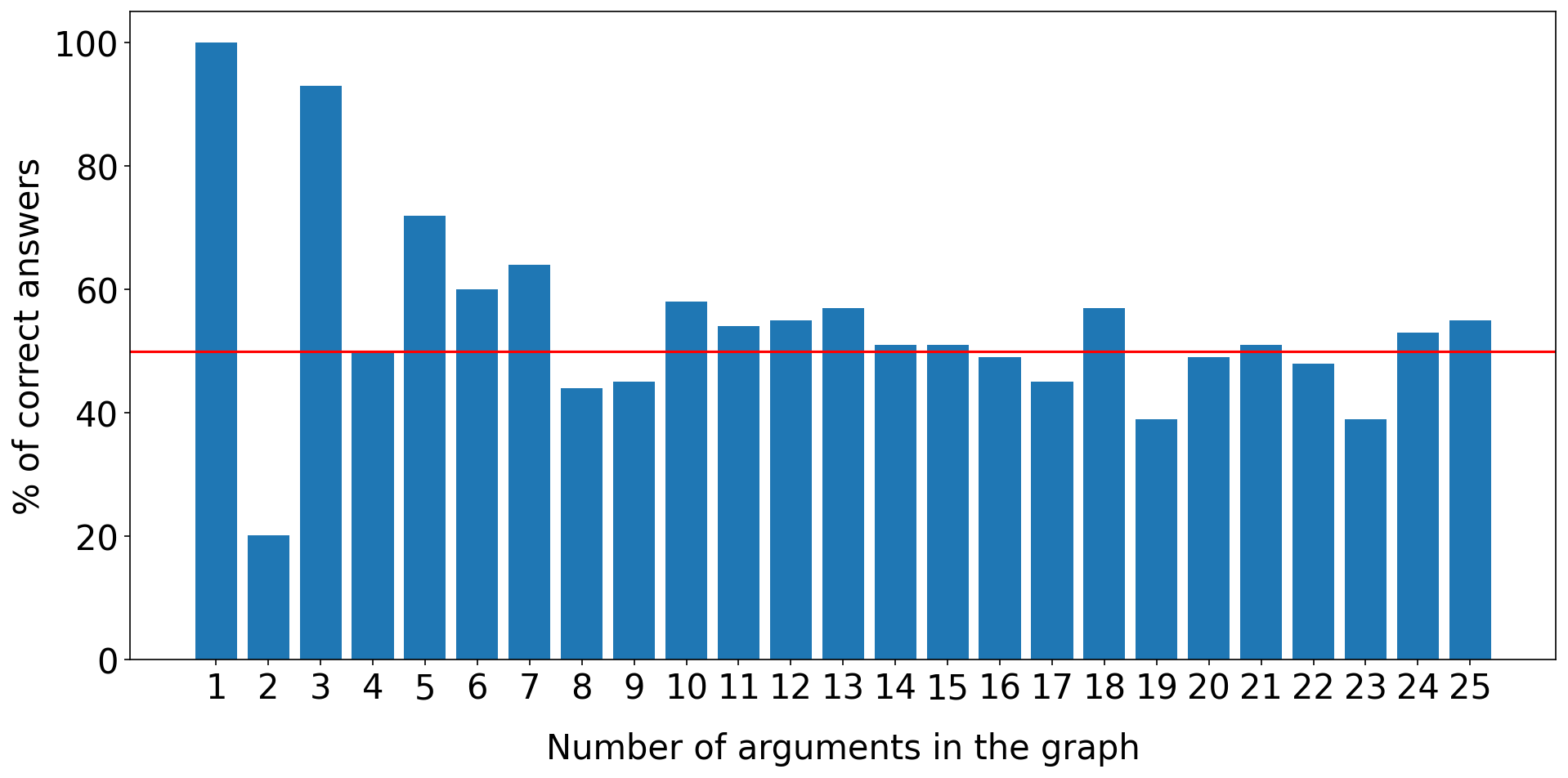}
         \caption{GPT-4o-mini}
         \label{}
     \end{subfigure}
    ~
     \begin{subfigure}[b]{0.49\textwidth}
         \centering
        \includegraphics[width = 1 \linewidth]{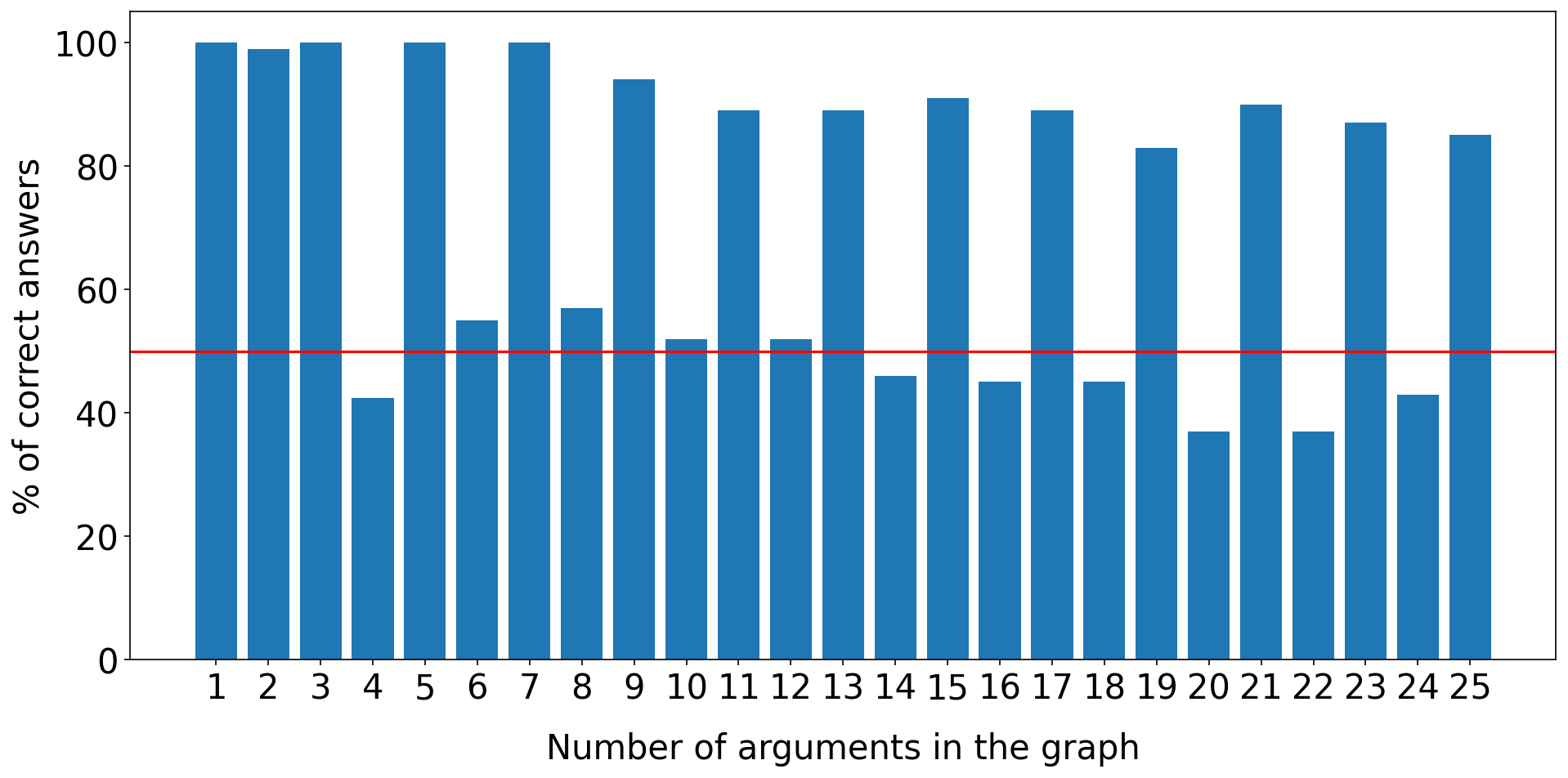}
         \caption{GPT-4o}
         \label{}
     \end{subfigure}
     \hfill
     \begin{subfigure}[b]{0.49\textwidth}
         \centering
        \includegraphics[width = 1 \linewidth]{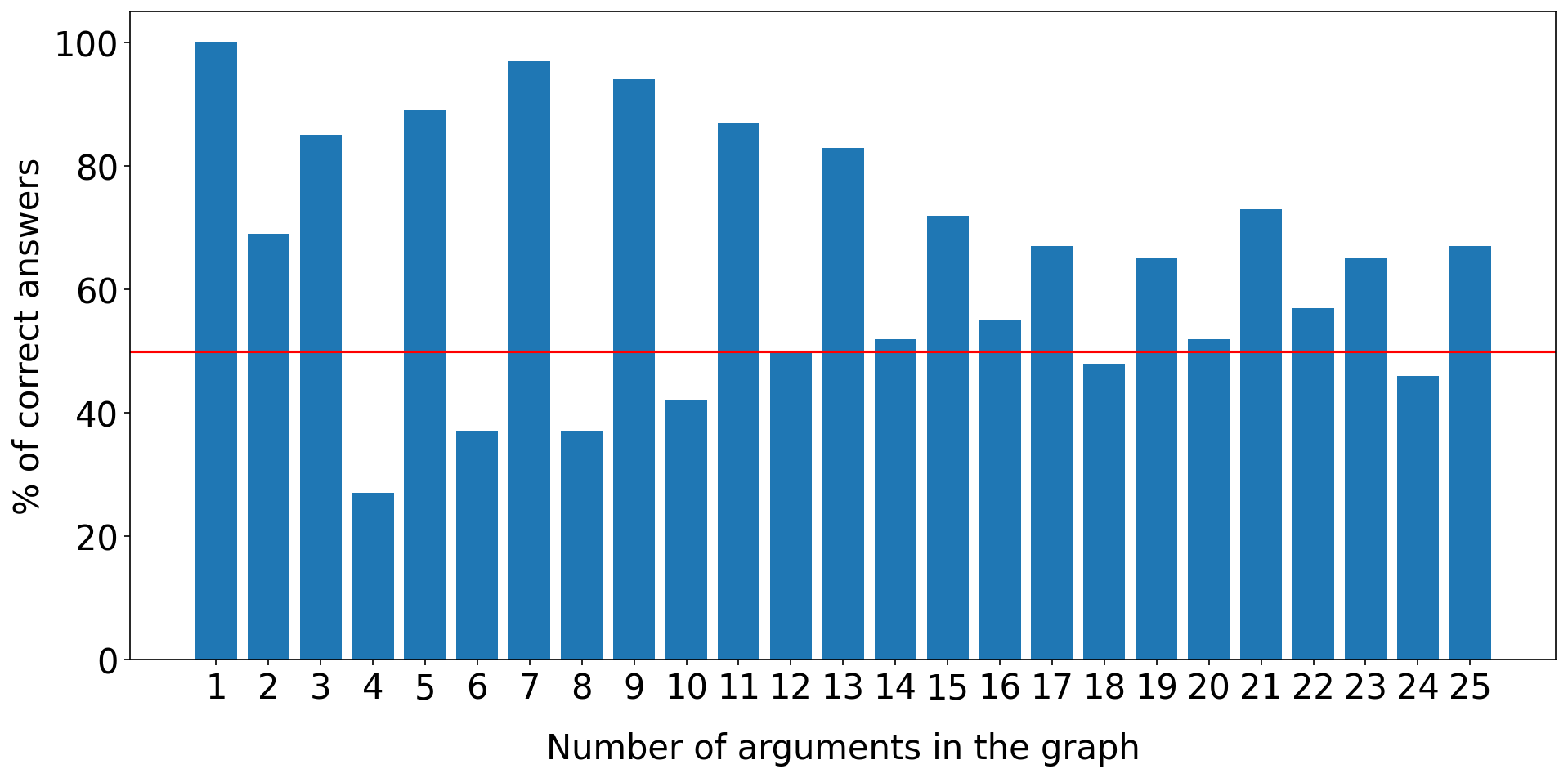}
         \caption{Gemini-1.5-flash}
         \label{}
     \end{subfigure}
    ~
     \begin{subfigure}[b]{0.49\textwidth}
         \centering
        \includegraphics[width = 1 \linewidth]{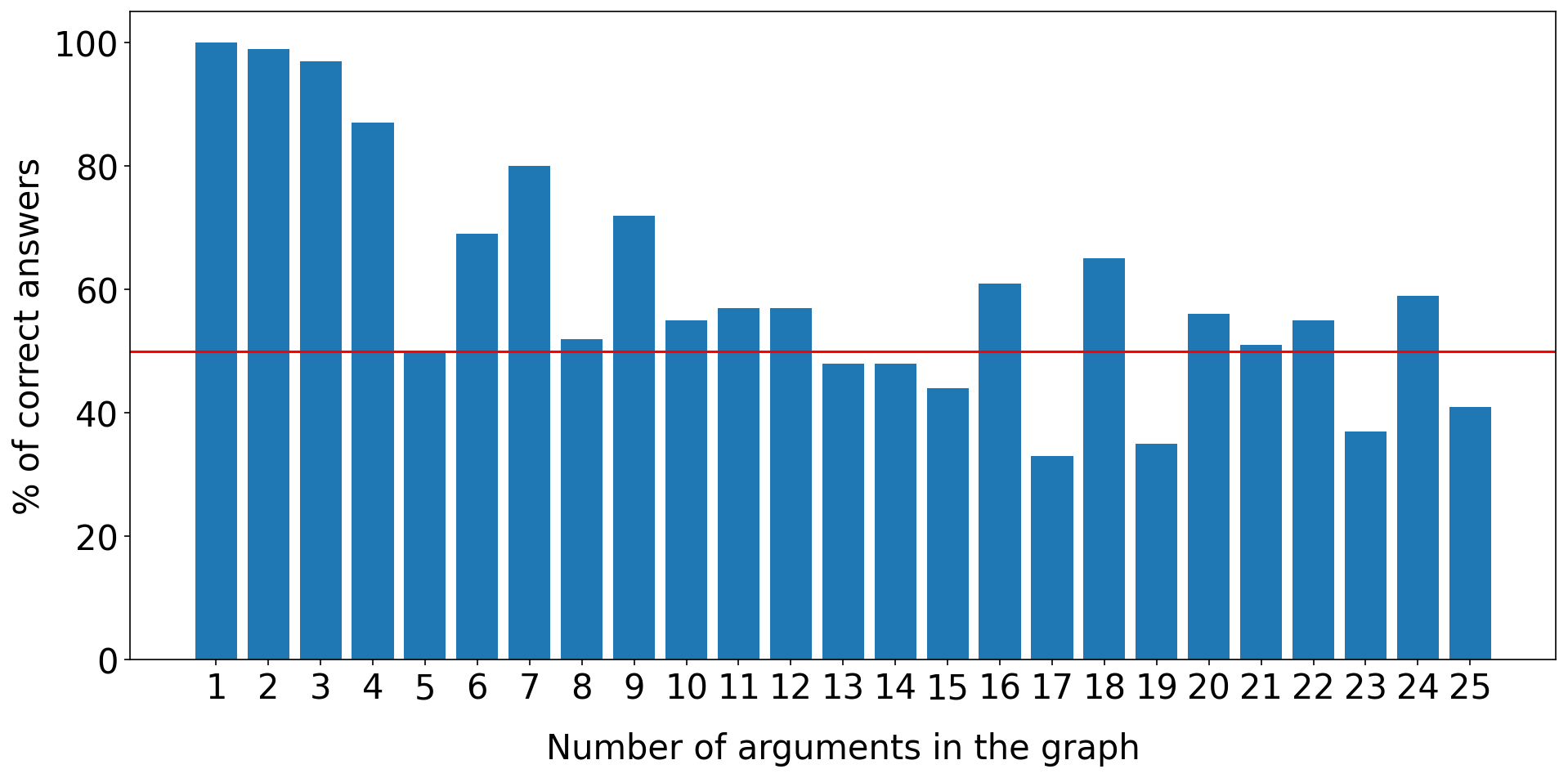}
         \caption{Gemini-1.5-pro}
         \label{}
     \end{subfigure}
     \hfill
      \begin{subfigure}[b]{0.49\textwidth}
         \centering
        \includegraphics[width = 1 \linewidth]{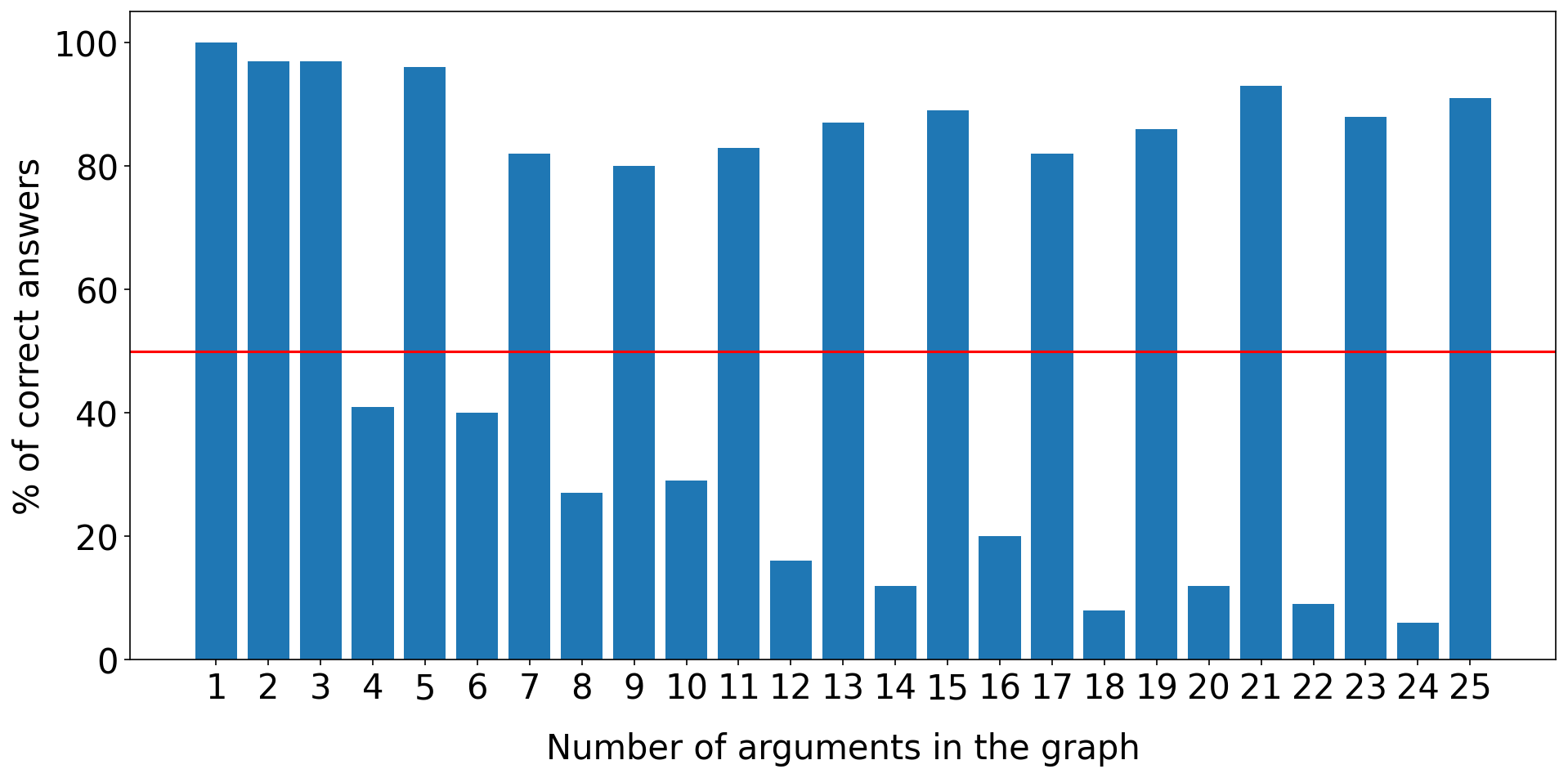}
         \caption{Claude-3-5-haiku}
         \label{}
     \end{subfigure}
    ~
     \begin{subfigure}[b]{0.49\textwidth}
         \centering
        \includegraphics[width = 1 \linewidth]{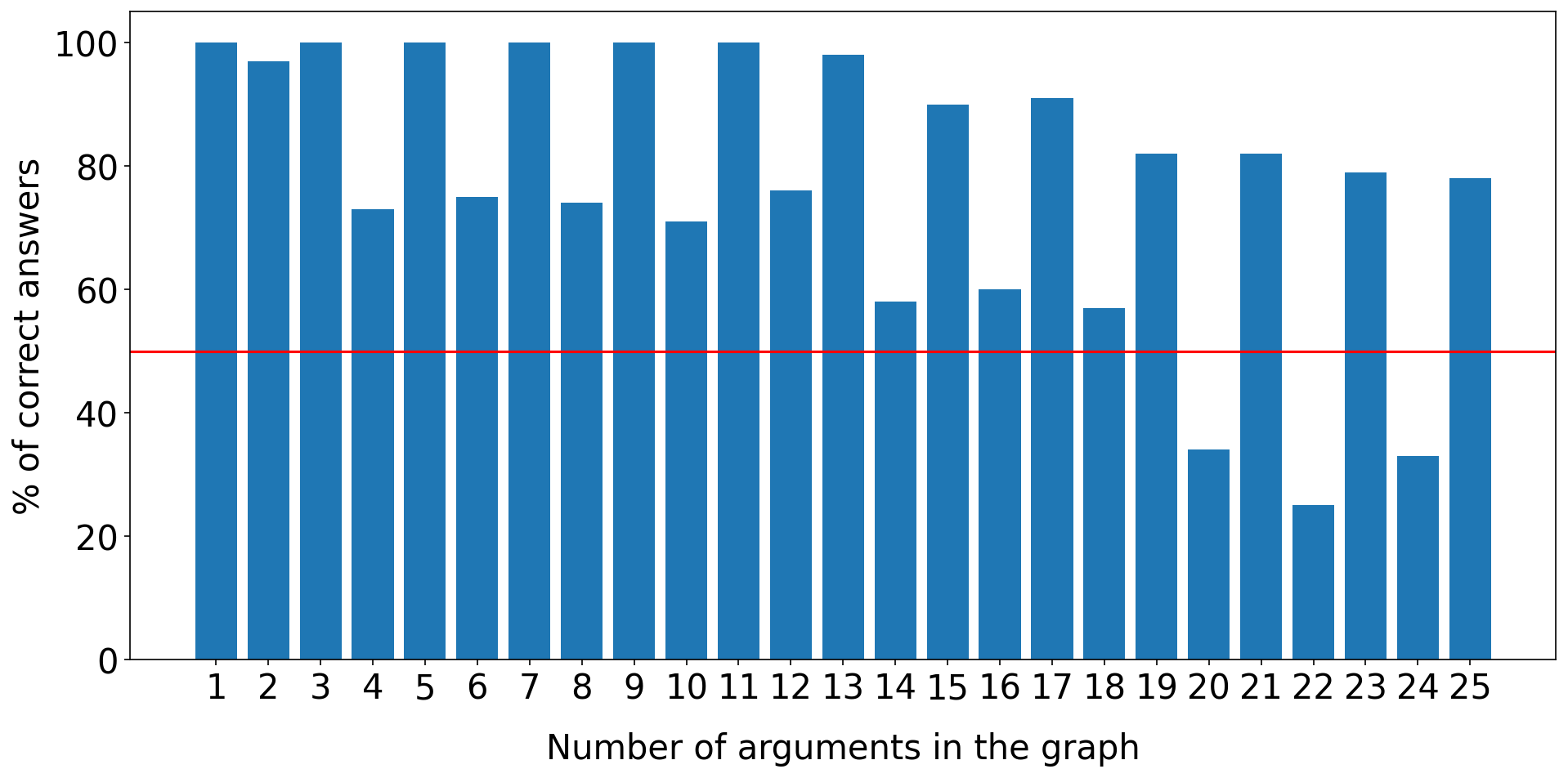}
         \caption{Claude-3-5-sonnet}
         \label{}
     \end{subfigure}
        \caption{Linear attack graphs: percentage of correct answers versus the number of arguments in the prompt}
    \label{fig:linear_per_num_args}
\end{figure*}

\begin{figure*}[h!]
     \centering
     \begin{subfigure}[b]{0.49\textwidth}
         \centering
        \includegraphics[width = 1 \linewidth]{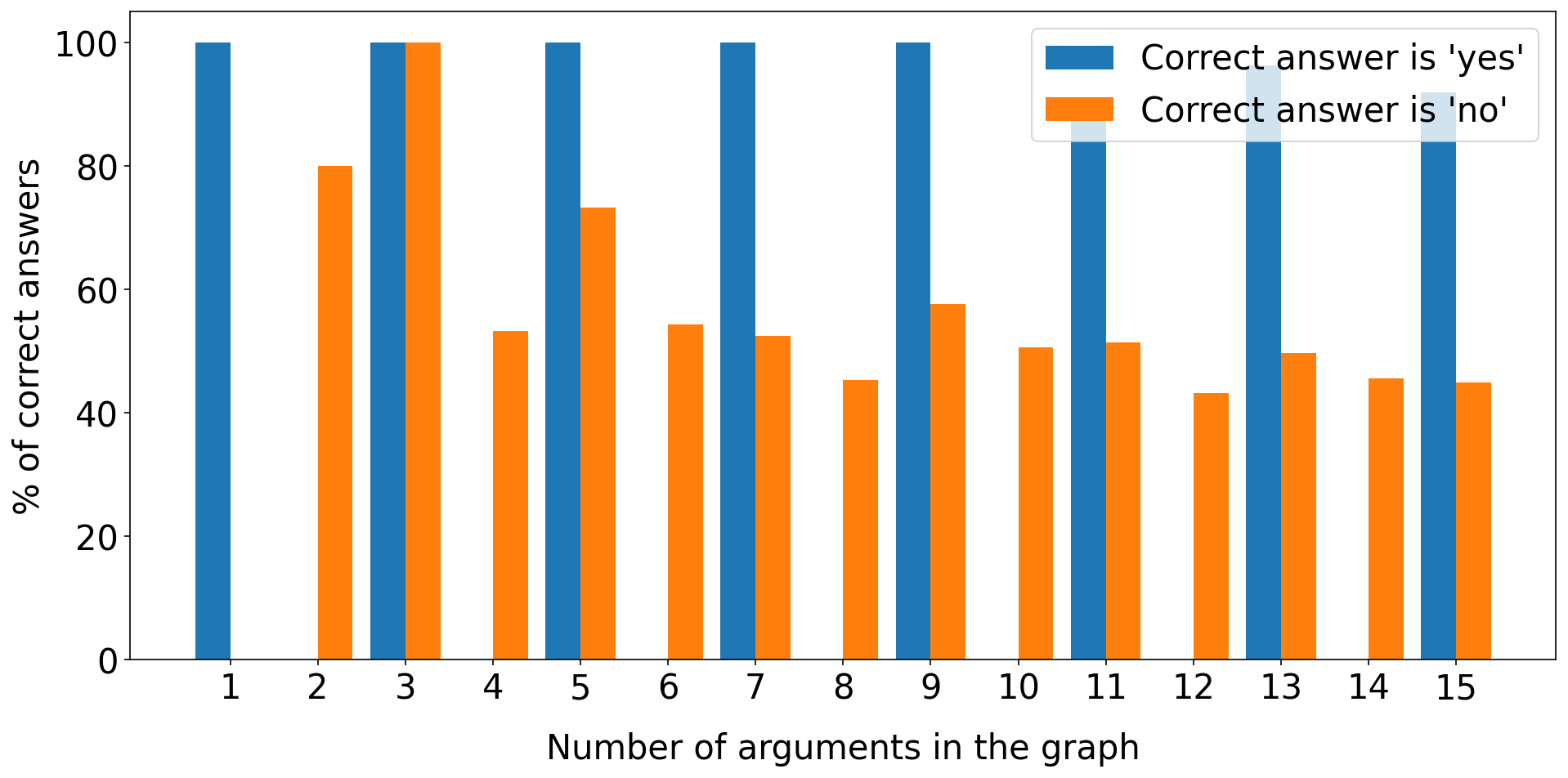}
         \caption{GPT-4o}
         \label{}
     \end{subfigure}
    ~
     \begin{subfigure}[b]{0.49\textwidth}
         \centering
        \includegraphics[width = 1 \linewidth]{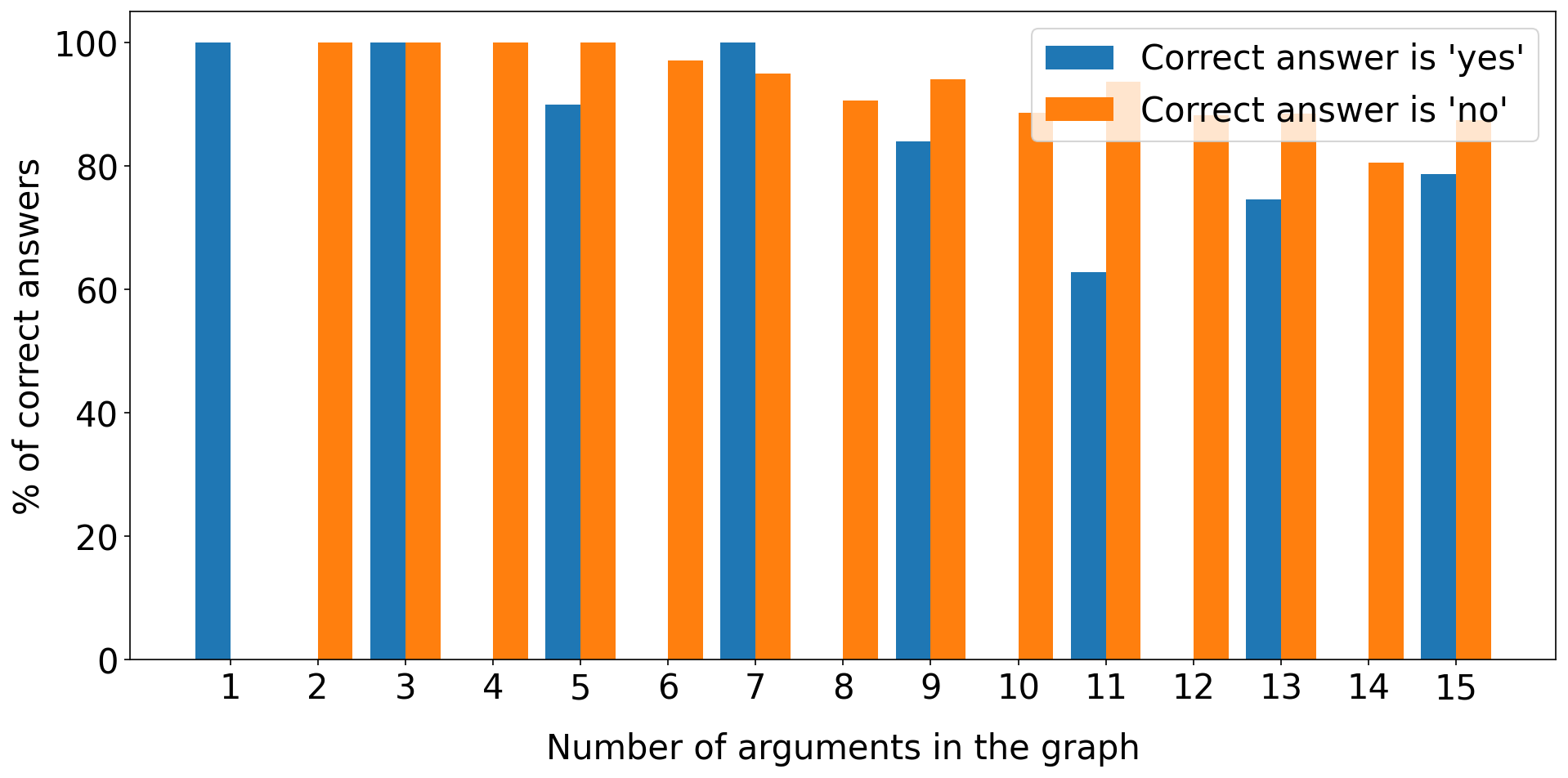}
         \caption{Claude-3-5-sonnet}
         \label{}
     \end{subfigure}
     \caption{Non-linear attack graphs: percentage of correct answers versus the number of arguments in the prompt, for prompts where the answer should be `yes' and where the answer should be `no'}
     \label{fig:nonlinear_per_num_args}
\end{figure*}

\begin{figure*}[h!]
     \centering
     \begin{subfigure}[b]{0.49\textwidth}
         \centering
        \includegraphics[width = 1 \linewidth]{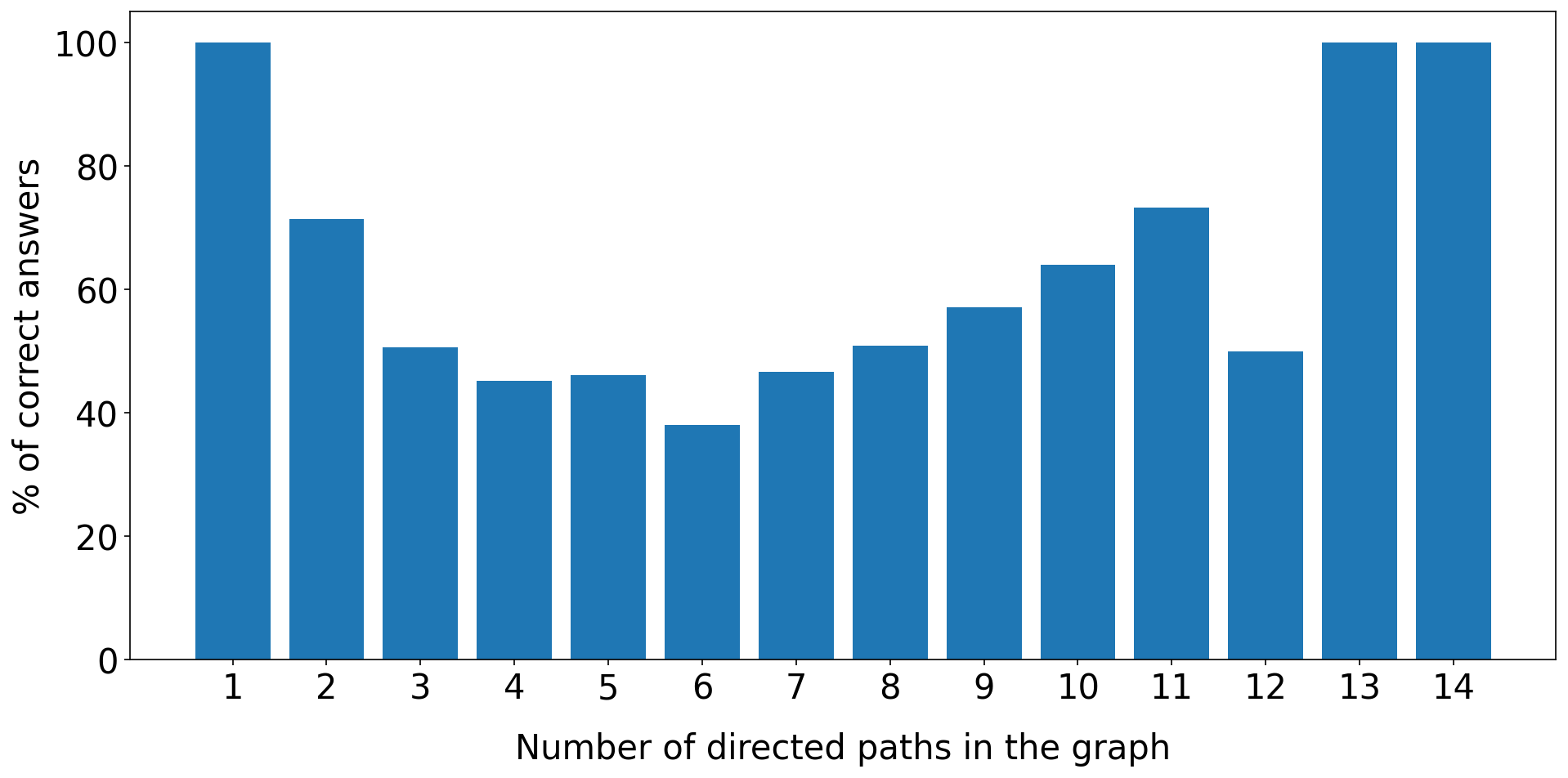}
         \caption{GPT-4o}
         \label{}
     \end{subfigure}
    ~
     \begin{subfigure}[b]{0.49\textwidth}
         \centering
        \includegraphics[width = 1 \linewidth]{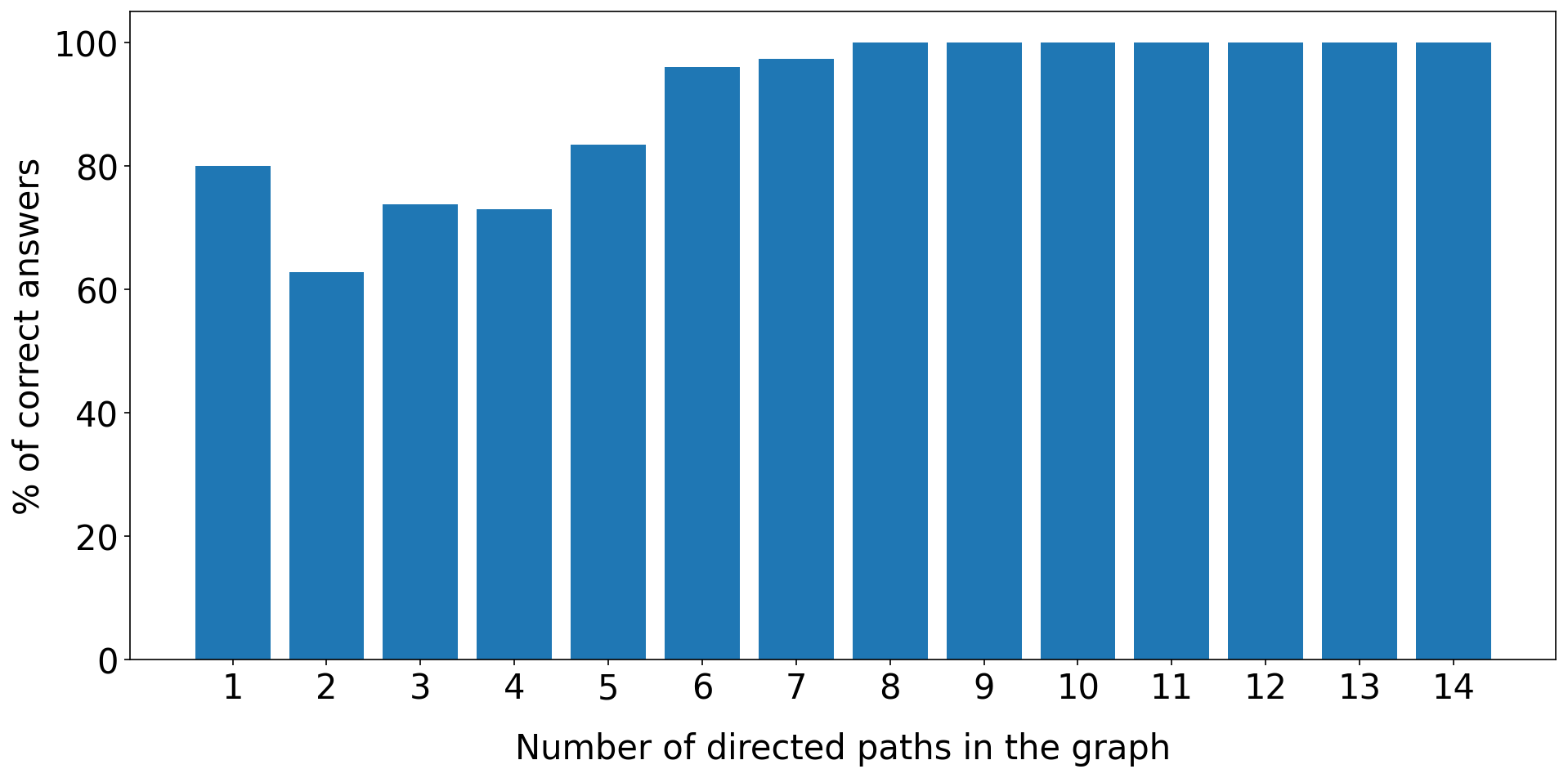}
         \caption{Claude-3-5-sonnet}
         \label{}
     \end{subfigure}
     \caption{Non-linear attack graphs: percentage of correct answers versus the number of directed paths in the prompt, based on prompts with a total of 15 arguments}
     \label{fig:nonlinear_per_num_chains}
\end{figure*}

\section{Results}
We show the mean accuracy, F1-score, Matthew's Correlation Coefficient (MCC), recall and precision for all seven models on the different benchmarks (Table~\ref{tbl:all_results}). Accuracy denotes the percentage of correct answers, which may invoke incorrect interpretations if the label distribution is skewed, as in our benchmarks. 
Recall describes how many prompts where the answer should be `yes' are correctly answered. 
Precision denotes how many prompts that were answered with `yes' by the model were answered correctly.
The commonly used F1-score combines recall and precision, but it does not account for true negative values. The MCC is the only metric that accounts for all four quadrants of the confusion matrix, and it is thus the metric that is most indicative of the overall performance. 
All values are normalized between 0 and 100, except for the MCC, which is normalized between -100 and 100. 
Each table also shows the experimental settings, which include the range of arguments used in the benchmark, the number of prompt variations, the total number of prompts in the dataset, and the number of outputs that were parseable into a binary `yes' or `no'. 

Table~\ref{tbl:linear_results} displays the performance of the models on prompts generated from the linear attack graphs. Most models are able to provide a parsable output for every prompt, except for the GPT models, which fail to provide a parsable output for two of the prompts. 
Because GPT-4o seemed to perform well for specific prompts with many arguments, we also tested GPT-4o with prompts that contain up to 50 arguments. Note again that we evaluated o1-preview with only the first 75 prompts. 

We show the performance of these models on prompts generated from non-linear attack graphs in Table~\ref{tbl:nonlinear_results}. Most of the prompts were parsable by the models. The performance of the models on the same prompts but with shuffled arguments is shown in Table~\ref{tbl:nonlinear_shuffled_results}. We also plot the mean MCC of each model for shuffled and non-shuffled prompts in Figure~\ref{fig:MCC_shuffled_diagram}. Note that the y-axis is scaled from 0 to 50.

Based on the results in Table~\ref{tbl:nonlinear_results}, Claude-3.5-sonnet and GPT-4o performed best on the non-linear prompts. We therefore collected all of the non-linear prompts where the Claude-3.5-sonnet and GPT-4o model made mistakes in order to create a set of hard prompts for the o1-preview model. For the experiment to remain feasible in terms of costs, we selected only the prompts from the first 508 unique prompts. Out of these 508 prompts, Claude-3.5-sonnet answered 56 prompts incorrectly and GPT-4o answered 236 incorrectly, with an overlap of 30 prompts. We selected these prompts and used them as input for the o1-preview model. The performance of o1-preview on these `hard' prompts can be seen in Table~\ref{tbl:nonlinear_o1}.

For the results on prompts based on linear attack graphs, we plot the number of correct answers versus the total number of arguments in the graph for all models in Figure~\ref{fig:linear_per_num_args}. We exclude the results from o1-preview, as it performed at 100\% accuracy for every number of arguments. For prompts based on non-linear attack graphs, we show the number of correct answers versus the total number of arguments in the graph in Figure~\ref{fig:nonlinear_per_num_args} for the two best performing models: GPT-4o and Claude-3.5-sonnet. In this figure, we split the performances based on whether the answer should be `yes' or `no'. Note that the correct answer should always be `no' for all prompts with an even number of arguments, and the answer is always `yes' for prompts with with one argument. For all other number of arguments, the answer can be either `yes' or `no'. 

Since these prompts are based on non-linear graphs, not only the total number of arguments, but also the number of directed paths and the number of arguments in each directed path play a role in the complexity of the overall graph and prompt. In Figure~\ref{fig:nonlinear_per_num_chains} we therefore also plot the performance of GPT-4o and Claude-3.5-sonnet versus the number of directed paths in the prompt, for all prompts with exactly 15 arguments (a total of 675 prompts). Here, we keep the number of arguments fixed to be able to interpret the potential connection between performance and the number of directed paths. 

We qualitatively evaluate the model output by selecting random prompts and comparing the outputs of the different models. To illustrate, the output of GPT-4o-mini and GPT-4o for the linear example prompt from Figure~\ref{fig:example_prompt} is shown in Figures~\ref{fig:example_answer_gpt4o-mini} and~\ref{fig:example_answer_gpt4o} respectively. GPT-4o-mini does not produce the right answer for this example prompt but GPT-4o does. In Figure~\ref{fig:example_answer_gpt4o_nonlinear}, we also show the output of GPT-4o for the non-linear example prompt shown in Figure~\ref{fig:example_prompt_nonlinear}, and in Figure~\ref{fig:example_answer_gpt4o_shuffled}, we show the output of GPT-4o on a shuffled version of the prompt shown in Figure~\ref{fig:example_prompt_nonlinear}. In these latter two outputs, GPT-4o produces the right answer for the shuffled version of the prompt, but produces the wrong answer for the non-shuffled version. 

\section{Discussion}
We discuss and reflect upon the results of the models on the three benchmarks: linear, non-linear and shuffled non-linear. 

\vspace{-10pt}

\subsection{Linear attack graphs}
In Table~\ref{tbl:linear_results}, we see the performances of the models on the prompts generated from linear attack graphs. The Claude-3.5-sonnet model from Anthropic performs best across all performance measures, followed by the GPT-4o model from OpenAI. 
A striking phenomenon can be observed in Figure~\ref{fig:linear_per_num_args}, where we plot the performance of the models against the number of arguments: generally speaking, most models seem to perform better on prompts with an odd number of arguments than on prompts with an even number of arguments (given that performance is higher than the 50\% baseline). For some models, there is an exception for prompts with two arguments, which GPT-4o, Gemini-pro, Claude-3.5-haiku, and Claude-3.5-sonnet can solve. Recall that all odd number of arguments result in a `yes' answer, whereas all even number of arguments result in a `no' answer. Most models therefore more often predict `yes' as an answer. Performance also generally seems to decrease with more arguments, and the performance of GPT-4o-mini, Gemini-flash, and Gemini-pro all approach the 50\% baseline performance after a certain number of arguments.

\begin{figure}[t]
    \footnotesize
    \raggedright
    \texttt{Based on the given information, we have two witnesses: Alice and Bob. Alice states that the train is late, which is a straightforward assertion. However, Bob claims that Alice is lying. \\ 
    Since we have no additional evidence or reason to believe Bob's statement of lying, we should initially give credence to Alice's testimony. That is, without evidence that Alice is indeed lying, we must accept his statement as true. \\
    Thus, the assertion that the train is late should be believed based on the current testimonies. \\ 
    Answer: yes.
    }
    \caption{Response of GPT-4o-mini to the prompt in Figure~\ref{fig:example_prompt}}
    \label{fig:example_answer_gpt4o-mini}
\end{figure}

\begin{figure}[t]
    \raggedright
    \footnotesize
    \texttt{To resolve this puzzle, we need to analyze the statements made by the witnesses and determine whether there is sufficient evidence to conclude whether the train is late. \\
    1. Witness Alice says that the train is late. \\
    2. Witness Bob says that witness Alice is lying. \\
    According to the rule that witnesses should be believed unless there is testimony that they are lying, we need to consider the impact of Bob's statement. Bob is explicitly saying that Alice is not telling the truth. However, the problem does not give us any explicit reason to distrust Bob, such as prior evidence or testimony stating she is lying. \\
    Therefore, according to the puzzle's instructions, Bob's statement that Alice is lying provides a reason not to believe Alice's claim. Since there's testimony (from Bob) that Alice might be lying, we cannot confidently say that the train is late based on Alice's statement.\\
    Answer: no}
    \caption{Response of GPT-4o to the prompt in Figure~\ref{fig:example_prompt}}
    \label{fig:example_answer_gpt4o}
\end{figure}

\begin{figure}
    \raggedright
    \footnotesize
    \texttt{To solve this puzzle, let's evaluate the statements and the implications of each witness's testimony. \\
1. Witness Alice says that the train is late. \\
2. Witness Bob claims that Alice lies. If Bob is telling the truth, then Alice is lying, and the train is not late. \\
3. Witness Charlie also claims that Alice lies. If Charlie is telling the truth, then Alice is lying, and the train is not late. \\
4. Witness Dan claims that Charlie lies. If Dan is telling the truth, then Charlie is lying, which means Alice is telling the truth, so the train is late. \\
If Alice's statement is true (and the train is late), it means Bob and Charlie are both lying. If Alice is lying (and the train is not late), it means Dan is also lying (because if Charlie is truthful, Alice lies, making Dan's statement false). \\
Therefore, Dan is the key to determining the truth. If Dan's statement is true, then Alice tells the truth and the train is late. If Dan's statement is false, then Charlie is telling the truth, Alice is lying, and the train is not late. \\
Since witnesses should be believed unless there is testimony they might lie, and there is no testimony contradicting Dan's truthfulness directly, the best reasonable assumption is to believe Dan. Thus, Alice is telling the truth, and the train is late.\\
Answer: yes}
    \caption{Response of GPT-4o to the prompt from Figure~\ref{fig:example_prompt_nonlinear}}
    \label{fig:example_answer_gpt4o_nonlinear}
\end{figure}

\begin{figure}
    \raggedright
    \footnotesize
    \texttt{To solve this reasoning puzzle, we need to analyze the credibility of each witness based on the claims they make about each other. \\
1. Dan's Statement: Dan claims that Charlie lies. We need to consider Dan's credibility. There is no testimony that Dan might lie, so we assume Dan tells the truth. \\
2. Charlie's Statement: Charlie claims that Alice lies. Since Dan (who we assume is truthful) claims that Charlie lies, Charlie is not credible.\\
3. Bob's Statement: Bob claims that Alice lies. There is no testimony against Bob, so we assume he tells the truth. \\
4. Alice's Statement: Alice claims that the train is late. However, both Charlie and Bob claim Alice lies. While Charlie's credibility is compromised, Bob's is not, meaning we should trust Bob's statement that Alice lies.\\
Given this analysis, Alice is likely lying based on Bob's testimony. Since Alice claims that the train is late, but we conclude Alice is lying, the train is not late.\\
Answer: no}
    \caption{Response of GPT-4o to a shuffled version of the prompt shown in Figure~\ref{fig:example_prompt_nonlinear}, where the order is: Dan, Charlie, Bob, and Alice}
    \label{fig:example_answer_gpt4o_shuffled}
\end{figure}

In~Figure~\ref{fig:linear_per_num_args}, we see that GPT-4o-mini, contrary to GPT-4o, scores poorly on prompts with two arguments. To understand why, we examine the output of the model based on the two argument example prompt in Figure~\ref{fig:example_prompt}, which can be seen in Figure~\ref{fig:example_answer_gpt4o-mini}. The produced output in this example claims that we should not believe Bob as there is no additional evidence to support his testimony that Alice is lying. However, this is in conflict with the rules of the prompt, which state that `Witnesses should be believed unless there is testimony that they are lying' (see Figure~\ref{fig:example_prompt}). In general, qualitative analyses show that models often state witness testimony should not be believed if there is an even number of arguments, and they justify their decision based on a lack of additional, supportive or corroborative evidence. Therefore, models tend to accept the first and main argument, even in situations where they should not. 

For prompts with an even number of arguments, better performing models, such as GPT-4o, tend to produce a similar type of output as GPT-4o-mini. The exception is for prompts with two arguments, which are almost always solved perfectly by the better performing models. To illustrate, the output for GPT-4o for the prompt in Figure~\ref{fig:example_prompt} can be seen in Figure~\ref{fig:example_answer_gpt4o}. This output is correct, as we cannot believe Alice's statement due to the witness testimony of Bob claiming that Alice is lying. While GPT-4o can perform the reasoning task well with two arguments, its performance is poor for all prompts with an even number of arguments of 4 or more (see Figure~\ref{fig:linear_per_num_args}), and it produces a similar type of justification as GPT-4o-mini does in Figure~\ref{fig:example_answer_gpt4o-mini}: it accepts the main argument and claims that there is no additional evidence that support the attacking arguments. 

While the models often fail to solve prompts with an even number of arguments, some models, such as GPT-4o, still perform relatively well on odd prompts with up to 25 arguments. We therefore also investigated the GPT-4o's performance with prompts up to 50 arguments using our approach. The overall performance on prompts with up to 50 arguments is lower than the performance of the model tested on prompts with up to 25 arguments, as can be seen in the second and last rows of Table~\ref{tbl:linear_results}. Generally speaking, the performance of GPT-4o thus decreases for prompts with more arguments, which was expected. The poor performance on the even number of arguments remains present, however.

While we only evaluated o1-preview with 75 prompts (3 iterations with 1-25 arguments), its performance was perfect, as shown in Table~\ref{tbl:linear_results}. The o1-preview model does not make the reasoning mistakes that GPT-4o or GPT-4o-mini make, but instead performs every task perfectly, regardless of the number of arguments. Examining the output of the o1-preview model reveals that the model occasionally identifies the odd-even pattern in the task and justifies its answer accordingly. 
This shows that it is possible for generative language models to solve the benchmark tasks as intended (for the relatively simple linear attack graphs).

\subsection{Non-linear attack graphs}
Table~\ref{tbl:nonlinear_results} shows the results of the models on prompts based on non-linear attack graphs. The Claude-3.5-sonnet model seems to be the best-performing model overall. The precision of GPT-4o, however, is higher than Claude-3.5-sonnet's on the non-linear prompts, as GPT-4o correctly answers almost all prompts where the correct answer is `yes'. Based on the F1-score and MCC, performance of the models is lower on non-linear prompts as compared to their performance on linear prompts, matching our expectations. 

In Figure~\ref{fig:nonlinear_per_num_args}, we show the percentage of correct answers for the best performing models on non-linear prompts versus the total number of arguments in the prompt. Note that in the linear benchmarks, only one unique attack graph can be made for any number of arguments. The answer for a given number of arguments is therefore always the same. This is not the case for non-linear arguments, where the correct answer is also dependent on the structure of the graph. Therefore, in Figure~\ref{fig:nonlinear_per_num_args}, we split the results based on whether the correct answer to the prompt is `yes' or `no'. Again, it should also be noted that the answer is always `no' for all even number of arguments, and the percentage of prompts where the answer is `yes' decreases as the total number of arguments increases.
As also observed from the precision and recall scores in Table~\ref{tbl:nonlinear_results}, we can see that GPT-4o tends to perform better on prompts where the answer should be `yes'. Note that GPT-4o also has such a high performance for `yes'-prompts on the linear attack graphs, as discussed earlier. In general, performance seems to decrease somewhat with more arguments across most models. 
The performance of the models does seem to be quite inconsistent, however, displaying the brittleness of the reasoning capabilities of these models.

Figure~\ref{fig:nonlinear_per_num_chains} shows the performance of the best performing models versus the number of directed paths. We show the results of the models on prompts with exactly 15 arguments, to better illustrate the effects. Here, the performance of GPT-4o first decreases and then increases again around six directed paths. A similar effect is observed for the Claude-3.5-sonnet model, where performance first decreases with two paths, and then increases to near 100\% accuracy. Based on these results, the best two models in this study therefore seem to perform best on prompts with few or many directed paths. With only one directed path, the graph becomes a linear attack graph with 15 argument, which both GPT-4o and Claude-3.5-sonnet can solve rather well (see Figure~\ref{fig:linear_per_num_args}). With relatively many directed paths, the paths become shorter and might be easier to interpret. Furthermore, the chance of a directed path of length 1 being attached to the main argument increases with the number of directed paths as well. Recall that a single directed path of 1 already makes the main argument unacceptable, which might be easy for the models to deduce. 

Figure~\ref{fig:example_answer_gpt4o_nonlinear} displays the response of GPT-4o to the example prompt from Figure~\ref{fig:example_prompt_nonlinear}. In this output, the model tries to reason from top to bottom: starting from Alice's statement and ending up at Dan's, ignoring Bob's statement in the process, and thus making a mistake. While Bob is mentioned by the model, his statement is not used in making the final deduction and instead only the longest directed path is used by the model. If GPT-4o generally reasons in this manner, its default answer will be `yes', as it presupposes the first argument it comes across. This seems to match the relatively high precision and low recall of GPT-4o in Table~\ref{tbl:nonlinear_results}.

In Table~\ref{tbl:nonlinear_o1}, we display the performance of the o1-preview model on the harder prompts. While the overall accuracy remains high, the o1-preview model does make mistakes when answering these non-linear prompts, unlike with the linear prompts as shown in Table~\ref{tbl:linear_results}. The precision remains at 100\%, as the o1-preview model makes no mistakes when the correct answer is `yes'. This shows that even language models specifically presented for reasoning capabilities can fail on relatively straightforward reasoning tasks. 

\subsection{Effect of shuffling}
To make the task more difficult, we shuffle the order of the arguments in all of the non-linear prompts. The performance of the models on these shuffled prompts is shown in Table~\ref{tbl:nonlinear_shuffled_results} and in Figure~\ref{fig:MCC_shuffled_diagram}. Interestingly, GPT-4o now has the highest performance in terms of MCC and recall for the shuffled prompts, whereas Claude-3.5-sonnet had the highest MCC and recall score for the non-shuffled prompts (see Table~\ref{tbl:nonlinear_results}. Even though the difference in MCC between the two models is small (0.10), the MCC of the GPT-4o model increases by 8.14 when compared to the non-shuffled results, whereas the MCC of the Claude-3.5-sonnet model decreases by 16.48. In Figure~\ref{fig:MCC_shuffled_diagram} we can see that for GPT-4o and Gemini-1.5-flash, the MCC increases after shuffling, and for Claude-3.5-haiku and Claude 3.5-sonnet the MCC decreases. For GPT-4o-mini and Gemini-1.5-pro the MCC remains almost the same. 

To understand why performance can increase for some models when arguments are shuffled, we can look at Figure~\ref{fig:example_answer_gpt4o_shuffled}, where we display GPT-4o's output for a shuffled version of the prompt shown in Figure~\ref{fig:example_prompt_nonlinear}, where we start with Dan's statement, then Charlie's, then Bob's, and finally Alice's statement. Just as for the non-shuffled prompt (Figure~\ref{fig:example_answer_gpt4o_nonlinear}), GPT-4o reasons from top to bottom, this time starting with Dan's statement and ending at Alice's statement. Because of this, the model is able to deduce the right answer, unlike with the non-shuffled prompt. This further seems to suggest that GPT-4o has a tendency of presupposing the first argument it reads. The behavior would also explain why for GPT-4o the recall is higher and the precision is lower on the shuffled prompts than on the non-shuffled prompts (see Table~\ref{tbl:nonlinear_results} and~\ref{tbl:nonlinear_shuffled_results}). GPT-4o does not produce the answer `yes' as often if the prompts are shuffled. Because there are more prompts for which the answer is `no' than `yes', the general performance (the MCC) of the model increases due to shuffling. 

\subsection{Effect of ontological variations}
In Figure~\ref{fig:linear_per_num_args}, we see that for most numbers of arguments, the models do not generate reasoning texts that are always correct nor always incorrect. Instead, models produce the correct output for some of the linear prompts, and the incorrect output for others, even if the number of arguments is the same. This difference can be caused by the fact that each prompt is unique, and is generated using the list of names and statements from the ontology. 
A similar effect can be found for the non-linear prompts as well, where a model might produce a correct output for a prompt based on an attack graph of a certain format, but an incorrect output for an ontological variation of the prompt based on the same attack graph. 
Perhaps some names or statements trigger a correct or incorrect response, even though such content specifics should not matter for the output. 
This illustrates the brittleness of the reasoning capabilities of generative language models, which is an undesirable trait for systems that should reason properly, such as those that would be used in the legal domain. 
A different explanation for the difference in output per number of arguments is that the generative language models are non-deterministic: they might produce an incorrect output when prompted once, and a correct output when prompted a second time. While we did not control for this non-deterministic nature of the models in this study, we argue that models in the legal domain should reason correctly all the time.

\section{Conclusion and future research}
We introduced an approach that combines attack graphs with an ontology to dynamically generate varied reasoning tasks of scaling complexity that can be used to evaluate the reasoning capabilities of generative language models. We illustrate our approach by generating a variety of prompts based on linear and non-linear attack graphs in combination with a list of names and statements. 

To demonstrate the viability of our approach, we evaluated seven commercially available language models on the benchmark. We chose these models as they are widely used and studied. 
Based on all of the results, we can draw a few conclusions. First of all, models perform worse on prompts that are based on a longer linear attack graph. Most models have difficulties with linear attack graphs of even lengths. This could be due to a bias towards accepting the first and main argument that is presented to the model, which would cause it to answer `yes', even though in all prompts with an even number of arguments the answer should be `no'. 

Prompts based on non-linear attack graphs also seem to be more difficult for models than those based on linear attack graphs: even the o1-preview model cannot solve all non-linear prompts, whereas it perfectly solves all linear prompts. 
Shuffling the arguments in the prompt does not necessarily make the task harder. Some models do perform worse with shuffled arguments, but others perform better. Shuffling could reduce the bias towards accepting the main argument, as the main argument is usually no longer shown first.
Performance overall is inconsistent and models provide different answers for the same attack graphs, illustrating the brittleness of their reasoning capabilities.

Our benchmark approach can be used to evaluate other generative language models as well, but evaluating their performance was beyond the scope of the current preliminary study. 
Considering the rapid developments in the field of language models, newer models may be able to solve the tasks that we generated in our experiments.
This is why our approach dynamically generates reasoning tasks with scaling complexity, such that more advanced models may be tested on more complex tasks. 

Our approach generates benchmarks to evaluate the performance of large language models on reasoning tasks. The performance is based solely on the evaluation of the output of the model. The exact reasoning patterns that lead to a given output are not investigated in this study. This is an important next step in the evaluation of the reasoning capabilities of generative language models, as it has been shown that machine learning models can learn to make the right decisions for the wrong reasons~\cite{StegingICAIL21}.

In future research, we wish to expand upon the approach by generating benchmarks for more diverse and complex reasoning tasks in the legal domain. In particular, we want to consider structured argument graphs, and employ other relationships between arguments. 
Moreover, we want to expand the ontology such that we can generate different attack and support relationships. By benchmarking for more realistic reasoning tasks, we can evaluate the legal reasoning capabilities of generative language models such as the ones used in the legal domain.

\begin{acks}
This research was funded by the Hybrid Intelligence Center, a 10-year programme
funded by the Dutch Ministry of Education, Culture and Science through the Netherlands
Organisation for Scientific Research, https://hybrid-intelligence-centre.nl.  
\end{acks}

\printbibliography

\end{document}